\documentclass[letterpaper]{article} 
\usepackage{aaai25}  
\usepackage{times}  
\usepackage{helvet}  
\usepackage{courier}  
\usepackage[hyphens]{url}  
\usepackage{graphicx} 
\usepackage{subfigure}
\usepackage{natbib}  
\usepackage{caption} 
\usepackage{booktabs} 
\usepackage{algorithm}
\usepackage{algorithmic}
\usepackage{amsfonts}
\usepackage{amsmath}
\usepackage{verbatim}
\usepackage{multirow}
\usepackage{multicol}
\usepackage{colortbl}
\usepackage{xcolor, xspace}
\usepackage{bm} 
\usepackage{newfloat}
\usepackage{listings}
\DeclareCaptionStyle{ruled}{labelfont=normalfont,labelsep=colon,strut=off} 
\floatstyle{ruled}
\newfloat{listing}{tb}{lst}{}
\floatname{listing}{Listing}

\newcommand\figref[1]{Fig.~\ref{#1}}

\newcommand\equref[1]{Eq.~(\ref{#1})}

\ifodd 1

\else

\fi

\newcommand{\sysname}{FoCTTA\xspace}
\title{
FoCTTA: Low-Memory Continual Test-Time Adaptation with Focus
}
\author {
    Youbing Hu\textsuperscript{\rm 1}, 
    Yun Cheng\textsuperscript{\rm 2}\thanks{Corresponding Author},
    Zimu Zhou\textsuperscript{\rm 3},
    Anqi Lu\textsuperscript{\rm 1},
    Zhiqiang Cao\textsuperscript{\rm 1},
    Zhijun Li\textsuperscript{\rm 1}\footnotemark[1]
}
\affiliations {
    \textsuperscript{\rm 1}Faculty of Computing, Harbin Institute of Technology\\
    \textsuperscript{\rm 2}Swiss Data Science Center, Zurich, Switzerland \\
    \textsuperscript{\rm 3}City University of Hong Kong\\
    youbing@stu.hit.edu.cn, yun.cheng@sdsc.ethz.ch, zimuzhou@cityu.edu.hk, \\luanqi@stu.hit.edu.cn zhiqiang\_cao@stu.hit.edu.cn, lizhijun\_os\@hit.edu.cn
}

\usepackage{bibentry}

\begin{document}

\maketitle

\begin{abstract}
Continual adaptation to domain shifts at test time (CTTA) is crucial for enhancing the intelligence of deep learning enabled IoT applications. However, prevailing TTA methods, which typically update all batch normalization (BN) layers, exhibit two memory inefficiencies. First, the reliance on BN layers for adaptation necessitates large batch sizes, leading to high memory usage. Second, updating all BN layers requires storing the activations of all BN layers for backpropagation, exacerbating the memory demand. Both factors lead to substantial memory costs, making existing solutions impractical for IoT devices. In this paper, we present FoCTTA, a low-memory CTTA strategy. The key is to automatically identify and adapt a few drift-sensitive representation layers, rather than blindly update all BN layers. The shift from BN to representation layers eliminates the need for large batch sizes. Also, by updating adaptation-critical layers only, FoCTTA avoids storing excessive activations. This focused adaptation approach ensures that FoCTTA is not only memory-efficient but also maintains effective adaptation. Evaluations show that FoCTTA improves the adaptation accuracy over the state-of-the-arts by 4.5\%, 4.9\%, and 14.8\% on CIFAR10-C, CIFAR100-C, and ImageNet-C under the same memory constraints. Across various batch sizes, FoCTTA reduces the memory usage by 3-fold on average, while improving the accuracy by 8.1\%, 3.6\%, and 0.2\%, respectively, on the three datasets. 

\end{abstract}

\section{Introduction}
After deploying deep neural networks (DNNs) to IoT devices for real-world applications, the model accuracy often severely degrades when there is a notable shift between the source and the target domain \cite{liang2020we,liu2021ttt++,ma2023letting,montesuma2023multi}.
In such cases, the pre-trained model should adapt to the target data distribution at test time without access to the target data labels, known as \textit{Test-Time Adaptation} (TTA) \cite{wang2020tent}.
As the target domain may evolve over time, continual test-time adaptation (CTTA) \cite{wang2022continual} is necessary, and has attracted increasing research interest \cite{chen2024towards,park2024layer,gao2024unified,wang2024continual}. 

As a continual unsupervised domain adaptation paradigm for IoT applications, CTTA must consider not only \textit{effectiveness} (e.g., accuracy), but also \textit{efficiency} (e.g., computation, memory, latency) when it comes to \textit{what}, \textit{when}, and \textit{how} to adapt.
Due to the unsupervised nature, TTA solutions \cite{wang2020tent,niu2023towards} often minimize an entropy-based loss by updating certain model parameters via standard mini-batch gradient descent.
The adaption can be performed upon all test-time samples or selectively to reliable samples.
For example, EATA \cite{niu2022efficient} selects reliable samples through entropy filtering for each predicted sample.
Most CTTA methods adopt a \textit{partial training} strategy, which only updates a small set of model parameters at test-time, for both effective adaptation and computation efficiency \cite{boudiaf2022parameter,park2024layer}.

\begin{figure}[t]
\centering
\includegraphics[width=1.0\columnwidth]{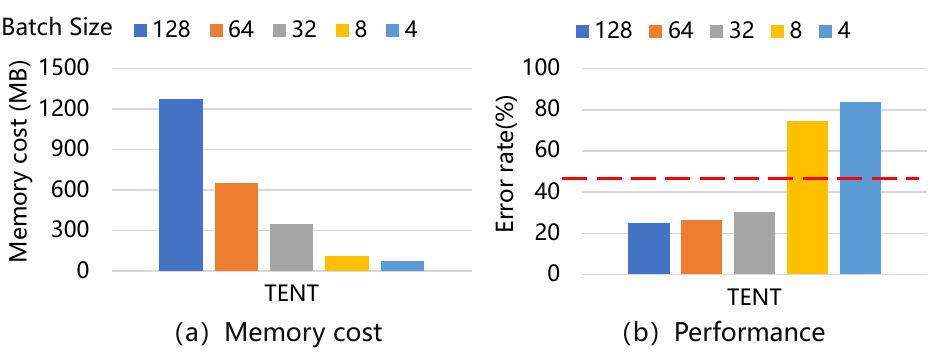}
\caption{Evaluate TENT memory cost and performance across various batch sizes.
(a) TENT memory costs at different batch sizes. (b) TENT performance at different batch sizes. The red dashed line represents the performance of the pre-trained model, i.e., without using any CTTA method. 
}
\label{fig:fig1}
\end{figure}

Although mainstream CTTA proposals \cite{niu2022efficient,yuan2023robust,niu2023towards,gong2023sotta} opt for \textit{computation} efficiency, it does not easily translate into \textit{memory} efficiency, which is crucial for IoT applications.
Specifically, they update the affine parameters of all batch normalization (BN) layers to adapt to the target domain. 
Even though a small set of parameters are updated, this strategy demands substantial memory to function properly.
\textit{(i)} 
A large batch size is necessary to accurately estimate the current batch statistics. 
\textit{(ii)} 
Updating the BN layers still involves storing activations for backpropagation, whose memory cost increases with the number of BN layers updated.
Both factors lead to considerable memory overhead to maintain high adaptation accuracy.
Fig.~\ref{fig:fig1}(a) shows the memory usage of TENT \cite{wang2020tent}, the earliest TTA method, with the increase of batch size.
As shown in Fig.~\ref{fig:fig1}(b), the BN-based method is sensitive to the batch size. 
In the target domain, it can only retain accuracy higher than the original pre-trained model (36.5MB) by consuming $8\times$ memory.

\begin{table*}[t]
    \centering
    \caption{\textbf{Comparing our approach and associated CTTA adaptation settings for memory efficiency.} Our method only optimizes the adaptation-critical representation layer, eliminating the need for large batch sizes and avoiding the storage of numerous activations. }
    \label{tab:tab1}
    \small
    \begin{tabular}{lcc}
        \hline

         Optimization parameters & Batch Size & Intermediate Activation \\
        \hline

        Parameters of all BN layers \cite{wang2020tent}    &  Large &    Large \\
        Parameters of specific BN layers \cite{hong2023mecta}  & Large & Smaller \\
        Additional prompt modules \cite{Song_2023_CVPR} &  Large &  Smaller \\
        \hline
        \textbf{Adaptation-critical representation layers (Ours)} &  Smaller &  Smaller\\
        \hline
    \end{tabular}
    \label{tab:1}
\end{table*}

A few pioneer CTTA schemes have been proposed to improve memory efficiency \cite{hong2023mecta,Song_2023_CVPR} (see Table~\ref{tab:tab1}).
Compared with standard methods that update all affine parameters in all BN layers \cite{wang2020tent,niu2022efficient}, these solutions either update shift-sensitive channels in the BN layers \cite{hong2023mecta,lim2022ttn}, or completely freeze the original model and only update extra side-way prompt modules \cite{Song_2023_CVPR,gan2023decorate}.
These schemes improve memory efficiency by reducing the number of adapted layers (channels), which saves the storage of activations during backpropagation.
However, they still rely on large batch sizes to boost adaptation accuracy, making them still sub-optimal for low-memory adaptation.

In this paper, we aim at memory-efficient CTTA that functions with small batch sizes and low activation storage during backpropagation. 
Specifically, we choose to update the representation layers rather than the BN layers to be more resilient to batch sizes. 
In addition, observing the importance of representation layers varies for adaptation, we only update the top-K critical representation layers to reduce the storage of activations.
By \textit{focusing on adaptation-critical representation layers}, we not only ensure memory efficiency but also achieve high adaptation accuracy.
This is implemented via an offline warm-up training stage after pre-training but before testing, where simulated unseen distribution shifts are used to identify shift-sensitive representation layers via a simple gradient-based importance metric.
At test time, only the top-K critical representation layers are updated for adaptation.
Experimental results show that our approach achieves state-of-the-art performance on standard benchmarks. 
Our main contributions are summarized as follows.
\begin{itemize}
    \item 
    We leverage the representation layers rather than the BN layers in the pre-trained model for CTTA.
    This paradigm shift mitigates the reliance on large batch size, an obstacle towards memory-efficient CTTA.
    \item 
    We empirically show that the importance of representation layers differs for TTA, and propose a simple metric to identify adaptation-critical representation layers.
    The layer-wise selective updating scheme notably reduces the storage of activations for TTA. 
    \item 
    We extensively validate the effectiveness of our solution on various models and datasets. 
    Under the same memory constraints, we outperform the state-of-the-art CTTA methods SAR, ECoTTA, and EATA,     showing accuracy improvements of 4.5\%, 4.9\%, and 14.8\% on CIFAR10-C, CIFAR100-C, and ImageNet-C, respectively.     Notably, we achieve a threefold reduction in average memory usage across different batch sizes, while boosting average accuracy by 8.1\%, 3.6\%, and 0.2\% on the three datasets. 
\end{itemize}

\section{Related Work}
\subsection{Test-Time Adaptation}
Test-time adaptation (TTA) addresses shifts between source and target domains during testing without accessing source data \cite{wang2020tent,lee2023towards,niu2023towards,park2024layer}. 
For example, TENT \cite{li2021test} proposes an entropy minimization based unsupervised test-time objective, and adapts to the target domain by updating the affine transformations in the BN layers. 
Alternatively, SHOT \cite{liang2020we} considers adaptation as source hypothesis transfer, and updates the feature representation layers to the target domain while keeping the classification layers unchanged.
As an orthogonal solution, EATA \cite{niu2022efficient} only selects reliable samples for adaptation. 
Continual TTA (CTTA) \cite{wang2022continual} extends the scope of TTA from a single target domain to a sequence of continuously changing domains. 
SWA \cite{yang2023exploring} explores safety supervision for CTTA. 
SAR \cite{niu2023towards} proposes a sharpness-aware and reliable entropy minimization to stabilize CTTA.
LAW \cite{park2024layer} leverages Fisher Information Matrix (FIM) to identify layers to keep or adapt.

Among the questions of \textit{what}, \textit{when}, and \textit{how} to adapt in CTTA, we focus on \textit{what} to update at low \textit{memory} cost, and adopt standard approaches for the other two issues, i.e., entropy-based sample selection \cite{niu2022efficient} and entropy minimization-based loss \cite{li2021test}.

\subsection{Efficient On-device Model Adaptation}
There is an increasing interest to enable model adaptation on memory-limited platforms \cite{lin2020mcunet}, where the bottleneck lies in the storage of activations and the use of large batch size for effective backpropagation \cite{cai2020tinytl}.
As a special model adaptation problem (unsupervised continual domain adaptation), CTTA faces the same memory bottleneck, and a few pioneer studies have explored memory-efficient CTTA.
For example, TENT \cite{wang2020tent} reduces activation during adaptation by exclusively updating BN layers.
MECTA \cite{hong2023mecta} further prunes activations of cached BN layers during backpropagation.
EcoTTA \cite{Song_2023_CVPR} keeps the entire model parameters frozen and only updates a small set of extra meta layers.
Although these approaches significantly reduce the memory consumption by saving storage of activations, they still depend on large batch sizes.
To operate with small batch sizes, SAR \cite{niu2023towards} replaces the BN with layer normalization (group normalization).
TTN \cite{lim2022ttn} adjusts the weight of the BN layers updated by the test-time batch according to the domain offset sensitivity of each BN layer.

Unlike the solutions above, we target at memory-efficient CTTA with both reduced activation size and batch size. 
The key idea is to update only a few representation layers sensitive to distribution shifts. 

\section{Problem Statement}
\label{sec_representational}

\paragraph{Continual TTA} 
We consider the standard continual test-time adaptation (CTTA) setup \cite{wang2020tent,liang2020we,wang2022continual}.
A model $f_{\bm{\theta}}(y|x)$ with parameters $\bm{\theta}$ is pre-trained on source data $D_s = (X^S, Y^S) = \{(x, y)\sim p_s(x,y)  \}$, where $x \in X^S$ is e.g., an image and $y \in Y^S$ is its associated label from the source class set $Y^S$. 
The target data are \textit{unlabeled} and sampled from an arbitrary target distribution $D_t = \{x \sim p_t(x)\}$, where $p_t(x)$ undergoes \textit{continual} changes over time $t$. 
Following the covariate shift assumption \cite{storkey2009training}, $p_s(y|x) = p_t(y|x)$ and $p_s(x) \neq p_t(x)$. 
As the target distribution gradually shifts from the source, $f_{\bm{\theta}}(y|x)$ 
no longer approximates $p(y|x)$, and needs adaptation to retain accuracy.
Unique in CTTA, $\bm{\theta}_t$ should make predictions online when source data is absent and adapt themselves into $\bm{\theta}_{t+1}$ for the subsequent input, given access to only $p_t(x)$ at time step $t$.

\paragraph{Memory Footprint of CTTA} 
In CTTA, the model $f_{\bm{\theta}}$ is often a neural network $f_{\bm{\theta}}(\cdot) = f_{\bm{\theta}_L}(f_{\bm{\theta}_{L-1}}(\cdots (f_{\bm{\theta}_1}(\cdot))\cdots))$, with parameters $\bm{\theta}_l$ in layer $l$. 
Assume the parameter $\bm{\theta}_l$ consists of weights $\bm{W}_l$ and bias $\bm{b}_l$, and the input and output features of this layer are $\bm{a}_l$ and $\bm{a}_{l+1}$, respectively. 
Given the forward pass $\bm{a}_{l+1} = \bm{a}_l\bm{W}_l + \bm{b}_l$, the corresponding backward pass with batch size 1 is
\begin{equation}\label{eq:1}
    \frac{\partial \mathcal{L}}{\partial \bm{a}_l} = \frac{\partial \mathcal{L}}{\partial \bm{a}_{l+1}}\bm{W}_l^T, \frac{\partial \mathcal{L}}{\partial \bm{W}_l} = \bm{a}_l^T\frac{\partial \mathcal{L}}{\partial \bm{a}_{l+1}},
    \frac{\partial \mathcal{L}} {\partial \bm{b}_l}=\frac{\partial \mathcal{L}} {\partial \bm{a}_{l+1}}
\end{equation}
Eq.~\ref{eq:1} shows that to update a learnable layer with weight $\bm{W}_l$, one must store all $\bm{a}_l$ to compute the gradients. 
Hence, for a model with $L$ layers, the memory cost during backpropagation given batch size $B$ can be estimated as
\begin{equation}\label{eq:2} 
    \bm{m}(cost) = \sum_{l=1}^L(\bm{m}(\bm{\theta}_l) + \bm{m}(\bm{a}_l) \cdot B)
\end{equation}
where $\bm{m}(\cdot)$ denotes the memory requirements. 
From Eq.~\ref{eq:2}, the memory cost of adaptation (via gradient descent) increases with \# layers $L$ updated and the batch size $B$.
Note that memory of the weights $\bm{\theta}$ are constant during adaptation.

\paragraph{Parameters to Update in CTTA}
Existing CTTA methods update the pre-trained model $f_{\bm{\theta}}(y|x)$ at test-time to better approximate $p(y|x)$.
The parameters $\bm{\theta}$ are often divided into adaptable weights $\bm{\theta}^a$ and frozen weights $\bm{\theta}^f$, where the adaptable weights are updated by minimizing an unsupervised loss $\mathcal{L}(x; \bm{\theta}^a \cup \bm{\theta}^f), x \sim p_t(x)$ w.r.t. $\bm{\theta}^a$ \cite{boudiaf2022parameter}. 
CTTA methods differ in their choices of the parameter partition $\{\bm{\theta}^a, \bm{\theta}^f\}$ and loss function $\mathcal{L}$, where the choice of adaptable weights $\bm{\theta}^a$ affects the memory cost.
Most CTTA proposals opt for \textit{computation efficiency}, which does not translate into \textit{memory efficiency} due to the storage of substantial activations and the use of large batch sizes.
Therefore, we focus on identifying adaptable weights $\bm{\theta}^a$ that yield low memory cost without compromising adaptation accuracy.

\section{Method}
\label{method}

We propose FoCTTA, a memory-efficient CTTA scheme that focuses on adaptation-critical representation layers.
FoCTTA updates representation layers rather than BN layers to mitigate dependence on a large batch size $B$.
It also selectively updates the representation layers to reduce the number of adaptable layers $L$.
From Eq.~\ref{eq:2}, the reduction of $B$ and $L$ decreases the memory usage during mini-batch gradient descent, the key memory bottleneck in CTTA.
Fig.~\ref{fig:3} illustrates the workflow of FoCTTA.

\begin{figure*}[ht]
	\centering
\subfigure[Gradient norm]{
    \includegraphics[width=0.23\textwidth]{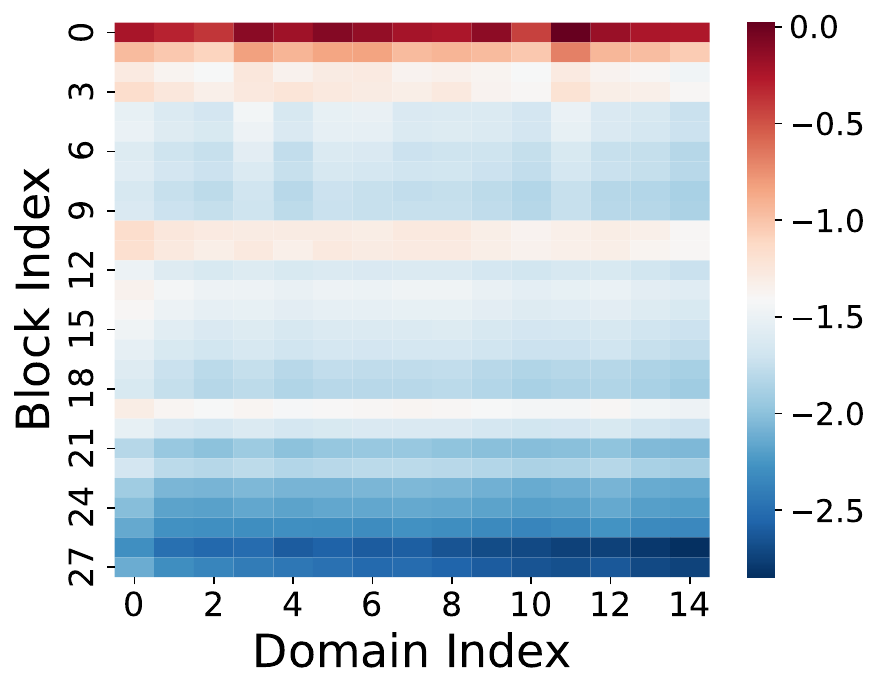}
}
   \subfigure[$\ell_1$ norm]{
    	\includegraphics[width=0.23\textwidth]{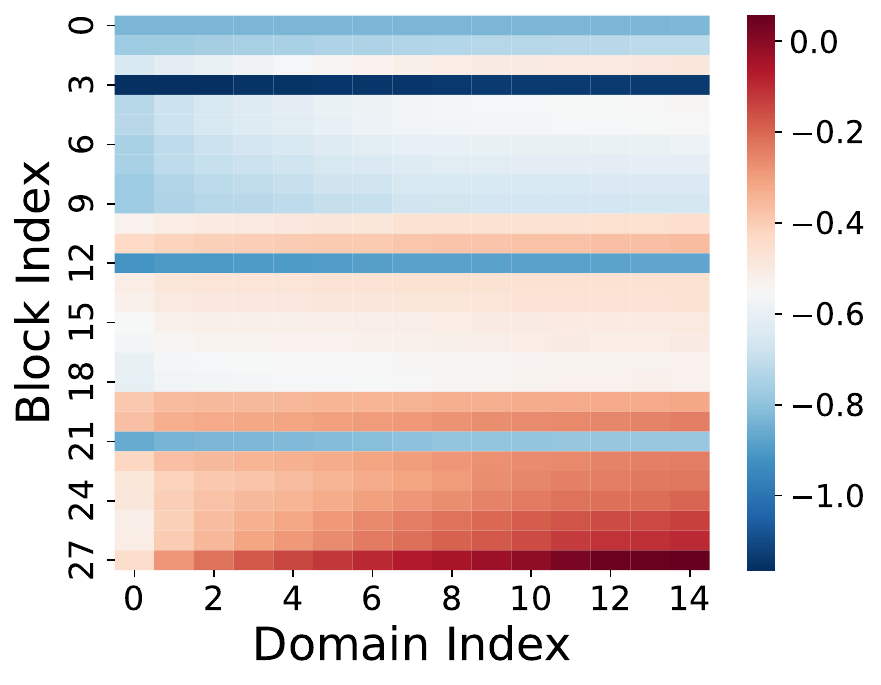}
    }
    \subfigure[Weight norm]{
    	\includegraphics[width=0.23\textwidth]{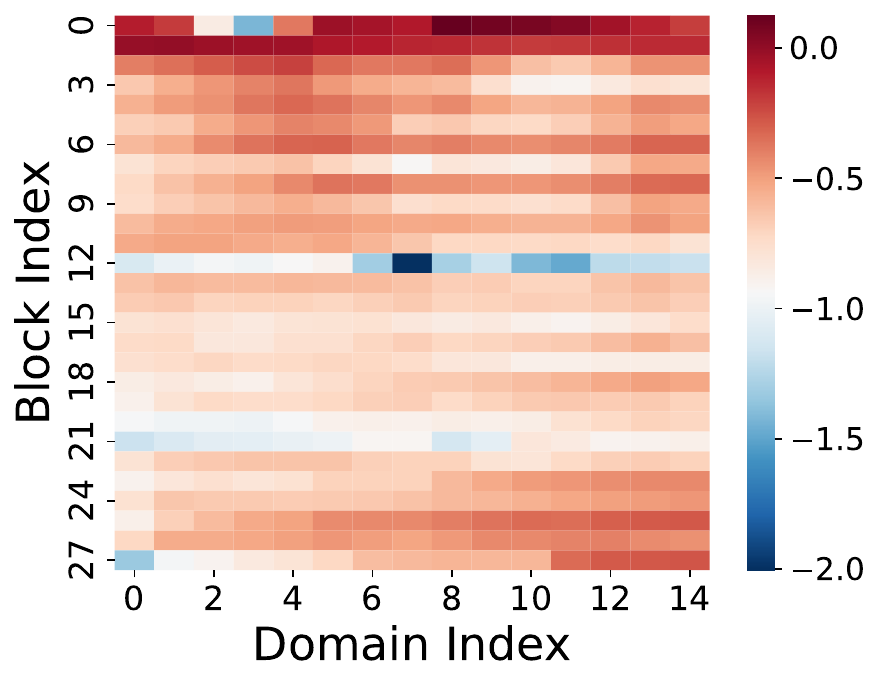}
    }
\subfigure[Performance]{
     \includegraphics[width=0.23\textwidth]{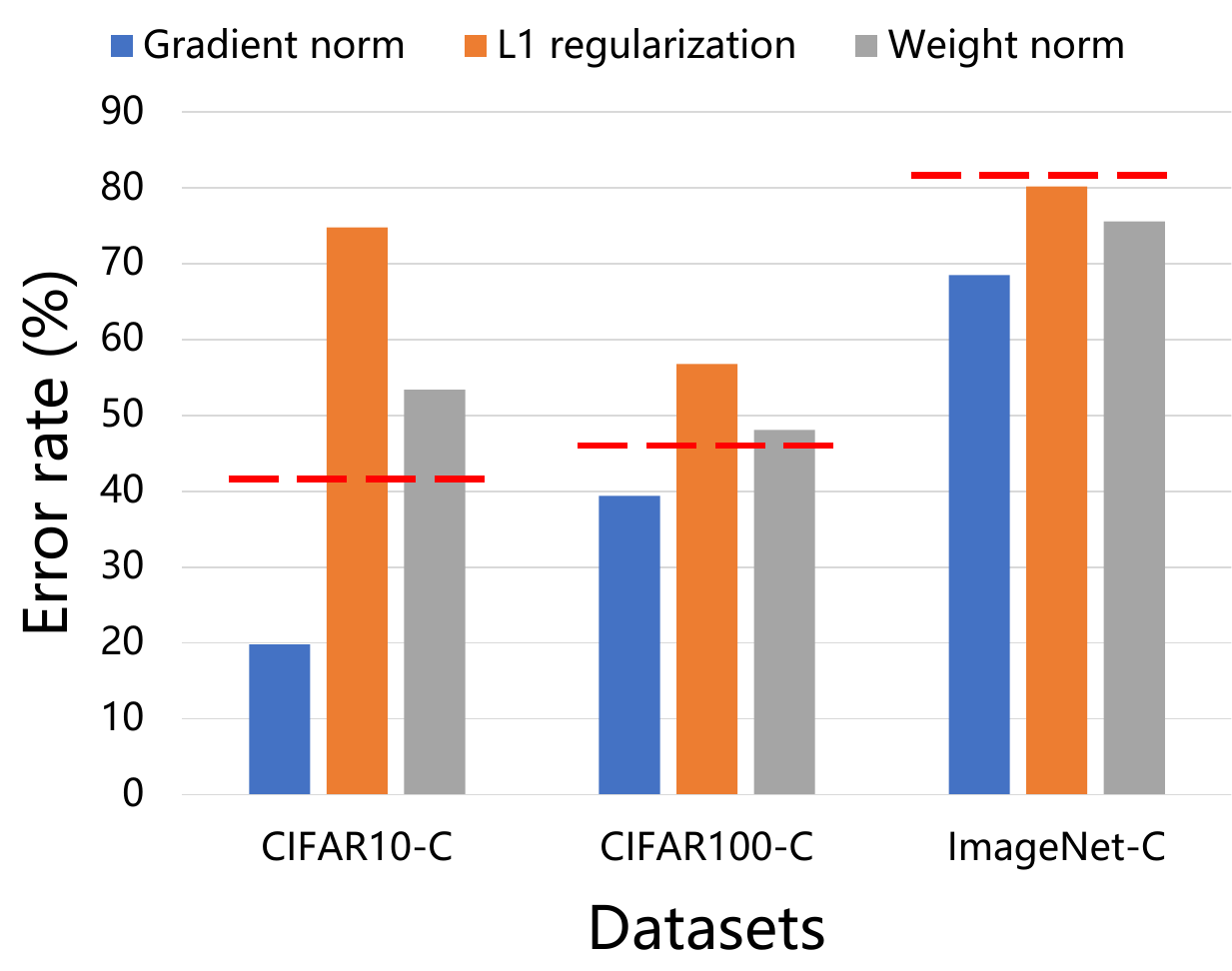}
}
\caption{\textbf{We evaluated the performance of selecting the top-K layers of the adaptation model using various metrics on three commonly used CTTA benchmarks. In addition, we also showcased the feature representation of CTTA during adaptation using various metrics on CIFAR10-C with WideResNet-28.} All these results are on a logarithmic scale, and we normalized them by linear transformation with the maximum value set to 0. A larger block index corresponds to deeper layers. \textbf{(a):} Gradient norm of different blocks. \textbf{(b):} $\ell_1$ norm of different blocks. \textbf{(c):} Weight norm of different blocks. \textbf{(d):} We optimized the top-K layers of the adaptation model on diverse datasets using varied metrics to assess its performance while keeping the other layers frozen. The red dashed line indicates the performance of the original model.
}
	\label{fig:2}
\end{figure*}

\subsection{Updating Representation Layers for CTTA}
As mentioned, a core issue in CTTA is to decide the adaptable parameters $\bm{\theta}^a$.
Inspired by the concept of source hypothesis transfer \cite{liang2020we}, we focus on representation layers for CTTA.
According to \cite{liang2020we}, $f_{\bm{\theta}}(y|x)$ can be decomposed into a feature extractor $g_s$ and a classifier $h_s$, where $f_{\bm{\theta}}(y|x) = h_s(g_s(y|x))$.
It suffices to update $g_s$ for CTTA.
Since adapting the feature extractor reduces the number of adaptable layers, it holds potential for memory-efficient CTTA.
Furthermore, as shown in \ref{performance_evaluation}, updating the representation layers works with small batch sizes, an essential advantage over BN layers.

We push the idea one step forward by asking the following:
can we achieve high CTTA accuracy by \textit{selectively} updating the representation layers?
That is, we hypothesize that the \textit{importance} of representation layers varies for CTTA, thereby only the critical ones needs updating.
Although layer importance and layer-wise fine-tuning have been extensively explored in network pruning \cite{li2016pruning,liu2017learning,lee2018snip,cheng2023survey}, their observations were intended for supervised learning in the same domain.
It is unclear whether similar observations and hypothesis hold for unsupervised adaptation to domains with shifts.

To this end, we conduct an empirical study to understand the importance of individual representation layers to CTTA.
Specifically, we pick three layer importance metrics: gradient norm \cite{molchanov2016pruning,lee2018snip}, $\ell_1$ norm \cite{liu2017learning,han2015learning}, and weight norm \cite{li2016pruning} commonly used for supervised training without domain shift, and select the top-K important layers ($K = 5$) for adaptation.
We tested various models and datasets in standard CTTA benchmarks.
Fig.~\ref{fig:2} shows the results.
More experimental details and results are shown in the supplementary material Sec.~A.
We make the following observations.
\begin{itemize}
    \item \textit{Importance of representation layers differs for CTTA}.
    The importance indicated by different metrics varies across layers.
    For instance, the gradient norm suggests important representation layers are the shallow layers, the $\ell_1$ norm shows deeper layers are more important, while the weight norm indicates the critical representation layers are distributed throughout the model.
    \item \textit{Gradient norm indicates layer importance for CTTA}.
    Across three datasets, we consistently observe that important representation layers selected by the gradient norm achieve the highest accuracy.
    The important layers identified by the other two metrics even yield performance lower than the model without adaptation.
\end{itemize}
These observations validate the hypothesis that layer importance also varies in the context of CTTA and show the gradient norm as an effective importance metric for CTTA.

\subsection{Identifying Critical Representation Layers}

\begin{figure}[t]
\centering
\includegraphics[width=\columnwidth]{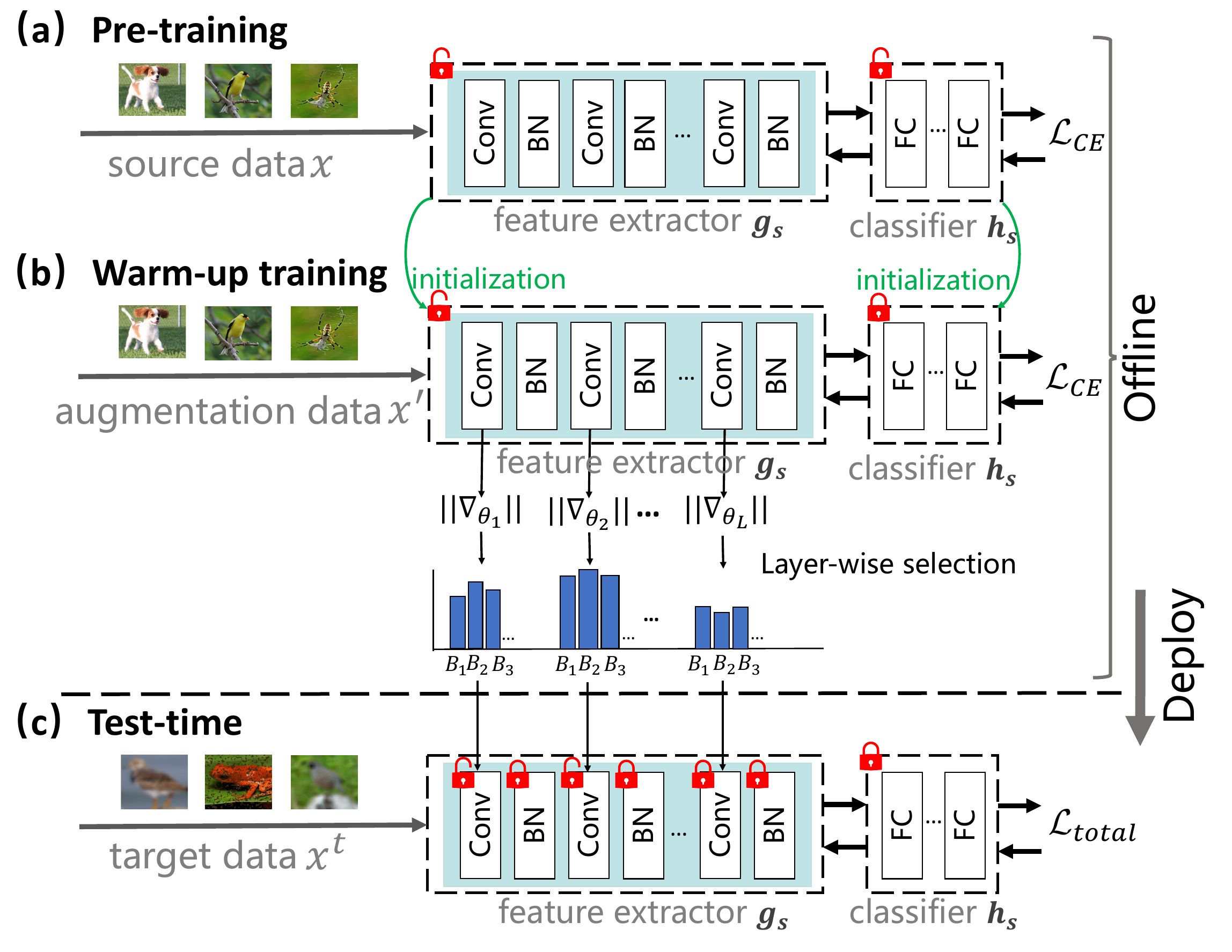}
\caption{
\textbf{The pipeline of our FoCTTA framework.}  \textbf{(a)} Pre-training, which is agnostic to architecture and pre-training methods, any pre-trained model can be used as initialization. \textbf{(b)} Warm-up training, which employs augmented data $x^\prime$ of the source data to simulate distributional shifts, computing the sensitivity of each layer in the feature extractor $g_s$ to domain shifts. \textbf{(c)} At test-time $t$, continuously changing target data $x^t$ is used as input, and only the adaptation-critical representation layer is optimized. 
}
\label{fig:3}
\end{figure}

To identify important representation layers sensitive to domain shifts, we leverage an additional \textit{warm-up training} phase after pre-training but before testing. 
It simulates domain shifts by augmenting the original training data.

Specifically, we take each original data point $x$ and create an augmented counterpart $x^\prime$ that shares the same semantic information. 
As shown in Fig.~\ref{fig:3}, in the warm-up training phase, we freeze the classifier $h_s$ of the pre-trained model, optimize the feature extractor $g_s$ with cross-entropy loss using the augmented data $x^\prime$ as input, and collect the gradient norms of each layer of $g_s$ in each batch\footnote{The warm-up phase shows robustness to different data augmentation types, as indicated by the ablation study results in Table~~\ref{tab:tab7}.}. 

The average gradient norm of each layer quantifies the layer's importance:
\begin{equation}\label{eq:3}
    \bm{s} = [\log {\frac{1}{B_N}\sum_{b=1}^{B_N} \Vert \nabla_{\theta_l}\Vert}]_{l=1}^{L}
\end{equation}
where $B_N$, $\theta_l$ and $\nabla_{\theta_l}$ are the $N$-th batch size, parameters of layer $l$, gradients of layer $l$, respectively. 
The vector $\bm{s}$ of length $L$ stores the importance of all layers in the pre-trained model. We then sort $\bm{s}$ from high to low.
The layers with the largest average gradient norms are identified as important for CTTA.
In practice, the selection is conducted by $\alpha \Vert \bm{s}\Vert$, where $\alpha$ is a tunable hyperparameter to balance memory cost and accuracy.
Note that the warm-up training is performed before test time, which is common in other CTTA solutions \cite{choi2022improving,jung2023cafa,lim2023ttn,niu2022efficient,Song_2023_CVPR}. 
Also, it does not require access to the source dataset $(X^S, Y^S)$ during test-time, and is agnostic to the architecture and pre-training method of the original model.

\subsection{CTTA Objective}
During test-time adaptation, FoCTTA exclusively optimizes adaptation-crucial layers for the target domains, maintaining the other layers unchanged. 
In line with \cite{niu2022efficient}, we utilize the entropy predicted by the adaptation model to identify reliable samples for subsequent model optimization. 
Consequently, the online adaptation loss function is formulated as
\begin{equation} \label{eq:4}
    \mathcal{L}_{ent} = \mathbb{I}_{\{H(\hat{y}) < H_0\}} \cdot H(\hat{y})
\end{equation}
\begin{equation} \label{eq:5}
    H(\hat{y}) = - \sum_{c}p(\hat{y})\log p(\hat{y}) 
\end{equation}
where $\hat{y}$ is the prediction output of a test image, and $p(\cdot)$ denotes the softmax function. 
The symbol $\mathbb{I}_{\{\cdot\}}$ represents an indicator function, and $H_0$ is a predefined hyperparameter.

In addition, to prevent catastrophic forgetting \cite{niu2022efficient,wang2022continual} and error accumulation \cite{tarvainen2017mean,arazo2020pseudo} due to long-term continuous adaptation, we add a regularization term to the loss function when optimizing Eq.~\ref{eq:4}. 
The final loss function is formulated as
\begin{equation}\label{eq:6}
    \mathcal{L}_{total} = \mathcal{L}_{ent} + \lambda\sum_{m=1}^{M}\Vert \tilde{x}_m - x_m \Vert_1
\end{equation}
where $\lambda$ is a positive scalar to control the ratio between two terms in the loss function. 
$M = \alpha \Vert \bm{s}\Vert$ denotes the number of layers to be updated. 
The terms $\tilde{x}_m$ and $x_m$ represent the $m$-th output of the adapted model and the original model, respectively. 
Our evaluations show that a small portion of (1.0\%) representation layers need to be updated.


\section{Experiments}
\label{experiments}

\subsection{Experimental Setup}

\textbf{Datasets.} We use CIFAR10, CIFAR100 \cite{krizhevsky2009learning}, and ImageNet \cite{deng2009imagenet} as the source domain datasets, while CIFAR10-C, CIFAR100-C, and ImageNet-C serve as the corresponding target domain datasets. These target domain datasets were originally designed to evaluate the robustness of classification networks \cite{hendrycks2019benchmarking}. Each target domain dataset contains 15 types of corruption with 5 severity levels. Following \cite{wang2022continual}, for each corruption, we use 10,000 images for both the CIFAR10-C and CIFAR100-C datasets, and 5,000 images for the ImageNet-C dataset.

\textbf{Implementation Detail.} All experiments are performed using the PyTorch framework \cite{paszke2019pytorch}. To ensure fair comparisons, we employ identical pre-trained models, specifically the WideResNet-28 and WideResNet-40 models \cite{zagoruyko2016wide} sourced from RobustBench \cite{croce2020robustbench}, as well as the ResNet-50 \cite{he2016deep} model from TTT++ \cite{liu2021ttt++}. During the warm-up training phase, data augmentation techniques such as color jittering, padding, random affine, center cropping, invert, and random horizontal flipping were applied to all source data. Subsequently, the pre-trained model is fine-tuned for one epoch using cross-entropy loss and the Adam optimizer with a learning rate of 0.00025 to identify adaptation-crucial layers. At test-time, similar to \cite{wang2022continual}, we employ the Adam optimizer with a learning rate of 0.001 for the CIFAR datasets, while for ImageNet, a learning rate of 0.00025 is utilized. In \equref{eq:4}, the entropy threshold $H_0$ is set to $0.4 \times \ln$$C$, where $C$ denotes the number of task classes. We empirically set $\lambda$ = 1 and $\alpha$ = 0.1.

\textbf{Evaluation setup.} We assess the effectiveness and memory efficiency of various methods through two configurations: 1) \textbf{Under memory constraints}, we test the error rates for each method. 2) \textbf{Under the same batch size}, we compute the error rates and memory consumption for each method, including the model parameters and the activation size.  
We demonstrate the memory efficiency of our work by using the official code provided by TinyTL \cite{cai2020tinytl}.

\begin{table*}[t]
\centering
\caption{\textbf{Comparison of error rate (\%) on the highest corruption severity under memory constraints}. We conduct experiments on CTTA setup.  CIFAR10-C uses WideResNet-28, CIFAR100-C employs WideResNet-40, and ImageNet-C utilizes ResNet-50. Avg. err means the average error rate (\%) of all 15 corruptions, and Mem. denotes total memory consumption, including model parameter sizes and activations. The lowest error is in bold, and the second lowest error is underlined.
}
\resizebox{\linewidth}{!}{
    \begin{tabular}{l|l|lllllllllllllll|cc}

        \toprule
         \multirow{2}{*}{Datasets}  &  \multirow{2}{*}{Method}  & \multicolumn{15}{l|}{t $\xrightarrow[\quad\quad
        \quad\quad\quad\quad\quad\quad\quad\quad\quad\quad\quad\quad\quad\quad\quad\quad\quad\quad\quad\quad\quad\quad\quad\quad\quad\quad\quad\quad\quad\quad\quad\quad\quad\quad\quad\quad\quad\quad\quad\quad\quad\quad\quad\quad\quad\quad\quad]{}$} & & \\
         &        & Gaus. & Shot & Impu.&  Defo.&  Glas.&  Moti. & Zoom&  Snow & Fros.&  Fog & Brig.&  Cont.&  Elas.&  Pixe.&  Jpeg&  Avg. err&  Mem. \\

        \hline
\multirow{9}{*}{\rotatebox{90}{CIFAR10-C}} 
     & EATA & \underline{24.5} & \textbf{20.2} &\underline{30.2}  &16.2 & \underline{32.9} & 19.9 & 16.9 & 22.1 & 22.4 & 19.2& 13.7 & 19.6 & 27.9 &22.6 &26.4 &22.3 &394.0  \\
     & Continual TENT & 25.1 & 22.8 & 35.4 & 21.9& 40.3 & 30.8 & 25.5 & 30.1 & 34.2 & 38.1 & 31.9 & 42.6 &44.7  &40.0 &46.6 & 34.0 & 394.0 \\
     & CoTTA & 99.8 & 99.8 & 99.8 &99.8 & 99.8&99.8 & 99.8 & 99.8 &99.8  & 99.8 & 99.8 & 99.8 & 99.8 & 99.8& 99.8 & 99.9 & 394.2 \\
    & SWA & 99.8 & 99.8 & 99.8 & 99.8& 99.8 & 99.8 & 99.8 & 99.8 &  99.8&99.8 & 99.8 & 99.8 &99.8 &99.8 & 99.8 & 99.9 & 394.2\\

    & ECoTTA & 27.8 & 22.8 & 31.3 &16.1 & 34.6 &17.8  &15.4  &19.9  & 18.7 & 17.6 & 12.1 & 16.5 & 27.1 & 22.2& 27.6 & 21.8 & 397.0  \\
    
    & SAR &29.0  & 26.9 & 35.7 & \underline{13.5}& 35.4 &\underline{14.9}  &\underline{13.1}  & \underline{18.4} & \underline{18.2} & \underline{15.8} & \underline{9.2} & \underline{13.7} & \underline{24.6} & \underline{20.6}& 28.8 &\underline{21.2} & 394.0  \\
     
     & LAW &26.6  &23.8  &31.2  & 22.0& 35.1 & 24.6 & 20.3 & 23.4 & 21.9 & 22.0 &16.6  & 19.1 & 27.2 & 22.8& \underline{24.6} & 24.1 & 378.8 \\

    &\cellcolor{lightgray}\sysname (Ours) &\cellcolor{lightgray}\textbf{22.4} &\cellcolor{lightgray}\underline{21.1} &\cellcolor{lightgray}\textbf{25.5} &\cellcolor{lightgray}\textbf{11.8} &\cellcolor{lightgray}\textbf{29.0} &\cellcolor{lightgray}\textbf{13.1} & \cellcolor{lightgray}\textbf{10.3}&\cellcolor{lightgray}\textbf{15.4}&\cellcolor{lightgray}\textbf{15.1} &\cellcolor{lightgray}\textbf{12.3} &\cellcolor{lightgray}\textbf{7.6}&\cellcolor{lightgray}\textbf{10.9} &\cellcolor{lightgray}\textbf{20.9}&\cellcolor{lightgray}\textbf{15.2}&\cellcolor{lightgray}\textbf{19.8}&\cellcolor{lightgray}\textbf{16.7} &\cellcolor{lightgray}356.0\\

       \hline
       \multirow{9}{*}{\rotatebox{90}{CIFAR100-C}} 
     & EATA & 45.4 & 43.5 & 47.0 &39.8 & 50.6 & 41.8 & 39.9 & 44.1 &44.0 &45.7  &38.3  & 40.8 &47.1  &43.0 &50.3  & 44.1 &47.9  \\
     & Continual TENT & 45.0 &48.3  &64.4  &78.5 &95.4  &96.5  &96.7  &96.9  &97.4  &97.6  &97.6  &98.0  &97.9  &97.7 &97.9  &87.0  & 47.9 \\
     & CoTTA & 64.7 &72.6  & 80.9 &85.1 & 91.7 & 94.2 & 96.4 & 97.2 & 97.3 & 97.9 &97.7  &98.2  &98.2  &98.3 & 98.4 &91.3  &50.7  \\
    & SWA & 62.3 & 71.8 & 79.8 & 83.6& 82.5 &93.7  & 96.2 & 97.1 & 97.3 & 98.0 & 97.6 & 97.7 & 98.0 &98.1 & 98.4 & 90.1 &50.7  \\
    & ECoTTA & \underline{44.9} & \underline{41.2} & \underline{44.9} &\underline{34.8} &\underline{45.5}  &\underline{35.5} & \underline{33.4} & \underline{38.5} &\underline{38.4}  &\underline{41.8}  & \underline{33.0} & \underline{38.9} & \underline{41.3} & \underline{36.3}& \underline{45.8} & \underline{39.6} &\underline{45.8}  \\
    & SAR &45.5  &44.9 & 49.5 &44.2 &56.4  &52.5  & 52.1 &62.2  &67.9  &80.3  &86.7  &96.5  &96.8  &97.2 & 97.2 &68.7  &47.9  \\
     & LAW & 61.7 &68.4  &83.8  &91.2 & 96.7 & 97.3 & 97.7 &97.8  & 98.0 & 97.9 & 97.8 & 98.1 &98.0  &98.0 &98.0  &92.0  & 46.1 \\
 
     &\cellcolor{lightgray}\sysname (Ours)  &\cellcolor{lightgray}\textbf{39.3} &\cellcolor{lightgray}\textbf{38.9}&\cellcolor{lightgray}\textbf{37.5}&\cellcolor{lightgray}\textbf{29.9} &\cellcolor{lightgray}\textbf{40.3} &\cellcolor{lightgray}\textbf{31.6} &\cellcolor{lightgray}\textbf{30.0} &\cellcolor{lightgray}\textbf{34.6}& \cellcolor{lightgray}\textbf{33.5}&\cellcolor{lightgray}\textbf{37.7} &\cellcolor{lightgray}\textbf{27.2} &\cellcolor{lightgray}\textbf{32.1}&\cellcolor{lightgray}\textbf{36.6} &\cellcolor{lightgray}\textbf{31.8}&\cellcolor{lightgray}\textbf{39.5}& \cellcolor{lightgray}\textbf{34.7} &\cellcolor{lightgray}45.9\\
       \hline
       \multirow{9}{*}{\rotatebox{90}{ImageNet-C}} 
     & EATA &92.7  & 91.3 & \underline{91.9} &\underline{92.9} & \underline{93.0} & \underline{85.5} & \underline{77.2} & \underline{76.5} & \underline{78.6} & \underline{67.0} & 51.5 & 88.4 & 71.8 &\underline{66.8} & 73.6 & \underline{79.9} &385.4  \\
     & Continual TENT & \underline{85.1} & \underline{86.8} & 94.6 &99.1 & 99.5 & 99.6 &99.6  &99.5  & 99.6 & 99.6 & 99.5 & 99.7 & 99.6 &99.6 & 99.6 & 97.4 & 385.4 \\
     & CoTTA & 96.0 &99.6 & 100.0 &100.0 &100.0  & 100.0 & 100.0 & 100.0 & 100.0 & 100.0 & 100.0 & 100.0 & 100.0 &100.0 & 100.0 & 99.7 &393.8  \\
    & SWA & 94.3 & 99.8 & 100.0 &100.0 &100.0  & 100.0 & 100.0 & 100.0 & 100.0 & 100.0 & 100.0 & 100.0 & 100.0 &100.0 & 100.0 & 99.6 &393.8  \\
    & ECoTTA & 94.8 & 97.9 & 100.0 &100.0 &99.9  & 100.0 & 99.9 & 100.0 & 100.0 & 99.9 & 100.0 & 100.0 & 100.0 &99.9 & 100.0 & 99.5  & 427.2 \\
    & SAR & 92.6 & 91.4 & 92.0 & 93.1& \underline{93.0} &\underline{85.5}  & 79.2 & 79.1 & 80.5 & 70.1 & \underline{51.3} & \underline{86.9} & \underline{71.3} & 67.6&\underline{73.0}  &80.4  & 385.4 \\
     & LAW  &90.7  & 96.6 & 99.6 &99.8 & 99.8 &99.7  & 99.7 & 99.8 & 99.8 & 99.7 & 99.8 & 99.8 & 99.8 &99.8 &99.8  &98.9  & 452.7 \\
 
     &\cellcolor{lightgray}\sysname (Ours) & \cellcolor{lightgray}\textbf{84.1}& \cellcolor{lightgray}\textbf{78.7} & \cellcolor{lightgray}\textbf{79.3}&\cellcolor{lightgray}\textbf{82.1} &\cellcolor{lightgray}\textbf{79.3} & \cellcolor{lightgray}\textbf{69.4} & \cellcolor{lightgray}\textbf{58.4} & \cellcolor{lightgray}\textbf{63.0} &\cellcolor{lightgray}\textbf{64.6}  &\cellcolor{lightgray}\textbf{50.4}  & \cellcolor{lightgray}\textbf{36.1} & \cellcolor{lightgray}\textbf{77.0} & \cellcolor{lightgray}\textbf{54.8} &\cellcolor{lightgray}\textbf{47.3} &\cellcolor{lightgray}\textbf{47.3}  & \cellcolor{lightgray}\textbf{65.1} &  \cellcolor{lightgray}423.0\\
       \bottomrule

    \end{tabular}
}

\label{tab:tab2}
\end{table*}

\begin{table*}
    \centering
    \caption{\textbf{Ablation experiments on CIFAR100-C with a batch size of 32.} ``Reg." refers to regularization, and ``LS." denotes layer selection. }
\resizebox{\linewidth}{!}{
    \begin{tabular}{l|lllllllllllllll|c}
        \hline
        Time & \multicolumn{15}{c|}{t $\xrightarrow[\quad\quad
        \quad\quad\quad\quad\quad\quad\quad\quad\quad\quad\quad\quad\quad\quad\quad\quad\quad\quad\quad\quad\quad\quad\quad\quad\quad\quad\quad\quad\quad\quad\quad\quad\quad\quad\quad\quad\quad\quad\quad\quad\quad\quad\quad\quad\quad]{}$}  &   \\
        \hline
        Method  & Gaus. & Shot & Impu.&  Defo.&  Glas.&  Moti. & Zoom&  Snow & Fros.&  Fog & Brig.&  Cont.&  Elas.&  Pixe.&  Jpeg&  Avg. err \\
        \hline
        \sysname &40.4 &39.8 &38.7 & 31.4&41.6 & 32.6 &31.5 & 34.8&34.9 & 38.3 &28.4 & 33.7& 38.2&32.7 &40.2 &35.8 \\
        w/o Reg. &40.5 &40.3 &41.5 & 32.6&41.6 & 32.6 &31.7 & 34.8&35.2 & 37.1 &28.9 & 33.6& 38.3&33.5 &39.8 &36.1 \\
        w/o LS. &99.0 &99.0 &99.1 & 99.0&99.0 & 95.8 &99.0 & 98.9&99.0 & 98.8 &99.0 & 99.0& 99.0&98.7 &99.0&98.7 \\
        w/o Reg. and LS. &98.6 &99.0 &99.0 & 99.0&98.4 & 97.1 &98.6 & 99.1&98.7 & 99.2 &98.4 & 98.7& 99.1& 98.8&99.0 &98.7 \\
        
        \hline
    \end{tabular}
}
    
    \label{tab:tab3}
\end{table*}

\textbf{Baselines.} We compare our method with several state-of-the-art continual test-time adaptation algorithms, includes 
Source, Continual TENT \cite{wang2020tent}, CoTTA \cite{wang2022continual}, ECoTTA \cite{Song_2023_CVPR}, EATA \cite{niu2022efficient}, SAR \cite{niu2023towards}, SWA \cite{yang2023exploring}, and LAW \cite{park2024layer}. All methods cannot access additional data in any way; i.e., these methods cannot be reset to the initial pre-trained model.

\subsection{Performance Evaluation}

\label{performance_evaluation}

\begin{table*}[t]
\centering
\caption{\textbf{Comparison of error rate (\%) and memory consumption (MB) on the highest corruption severity under the same batch size}. We conduct experiments on CTTA setup. CIFAR10-C uses WideResNet-28, CIFAR100-C employs WideResNet-40, and ImageNet-C utilizes ResNet-50. Err. means the average error rate (\%) of all 15 corruptions, Avg. err, and Avg. mem means the average error rate (\%) and the average memory (MB) of overall batch sizes, respectively. Mem. denotes total memory consumption, including model parameter sizes and activations. The lowest error is in bold, and the second lowest error is underlined.
    }
\resizebox{\linewidth}{!}{
    \begin{tabular}{l|l|llllllllllll|cc}
    
     \toprule
      \multirow{3}{*}{Datasets} & \multirow{3}{*}{Method} & \multicolumn{12}{c|}{Batch Size} \\
         & & \multicolumn{2}{c}{128}  & \multicolumn{2}{c}{64} & \multicolumn{2}{c}{32} & \multicolumn{2}{c}{16} & \multicolumn{2}{c}{8} & \multicolumn{2}{c|}{4} &   \\
         & & Err.  & Mem.   & Err.  & Mem.  & Err.  & Mem.   & Err.  & Mem. & Err.  & Mem.   & Err.  & Mem. & Avg. err & Avg. mem\\
        \hline
       \multirow{9}{*}{CIFAR10-C} & Source & 43.5 &204.3 &43.5  &120.4 &43.5 &78.4 &43.5 &57.5 &43.5 &47.0 & \underline{43.5} &41.7 &43.5 & 91.6\\
       & EATA & 18.7 &1240.3 & 20.3 &638.4 & 23.9&337.5 & 28.9 &187.0 & 47.2 &111.8 & 47.2 &74.2& \underline{31.0} & 431.5\\
        &Continual TENT &25.6  &1240.3 & 38.1 &638.4 & 46.5& 337.5&62.2&187.0 &78.7&111.8&86.4&74.2&56.3&431.5 \\
        &CoTTA &17.9 &2939.1&18.7&1688.4&22.2 & 1063.1& 34.5&750.4 & 59.3 &594.1 & 79.0 &515.9 & 38.6&1133.4\\
        &SWA & 17.7 & 2939.1& 18.6 &1688.4 &21.7 &1063.1 &31.8 &750.4 & 56.9 &594.1 & 76.3 &515.9 &37.2&1133.4 \\
        &ECoTTA & 19.6 &747.2 & 21.8 & 397.0 &22.3 &221.8 &39.8 &134.3 & 46.7& 90.5& 54.2&68.6 & 34.1&276.6\\
        &SAR & 20.4 &1240.3 &20.7&638.4 & 21.4&337.5 & 22.9&187.0 & \underline{32.6}&111.8& 75.7 &74.2 &32.3 &431.5\\
        &LAW &\textbf{16.3} &2647.3& \underline{17.5} & 1396.6&\textbf{18.6}&771.3 &\underline{22.8}&458.6 &51.2&302.3 &78.2 &224.1 &34.1&966.7\\
        
        &\cellcolor{lightgray}\sysname (Ours) & \cellcolor{lightgray}\underline{16.7} &\cellcolor{lightgray}356.0 &\cellcolor{lightgray}\textbf{17.3}&\cellcolor{lightgray}197.9 &\cellcolor{lightgray}\underline{18.9}& \cellcolor{lightgray}118.9&\cellcolor{lightgray}\textbf{21.6}& \cellcolor{lightgray}79.4& \cellcolor{lightgray}\textbf{27.1} & \cellcolor{lightgray}59.7 &\cellcolor{lightgray}\textbf{35.7}&\cellcolor{lightgray}49.8 &\cellcolor{lightgray}\textbf{22.9}&\cellcolor{lightgray}143.6\\
        \hline

        \multirow{9}{*}{CIFAR100-C} 
        & Source  &46.8&35.8 & 46.8&19.0&46.8& 10.6& 46.8&6.4 & \underline{46.8}& 4.4& \textbf{46.8}&3.3 & \underline{46.8} &13.3\\
       & EATA  & 36.1 &367.2 & 37.0& 184.7& 39.7 &93.5 &44.1 & 47.9& 51.7&25.1 & 74.7 &13.7 & 47.2 &122.0\\
        &Continual TENT  &41.3&367.2& 49.0 &184.7&79.2 &93.5 &87.0 &47.9 &95.4 &25.1 &98.3& 13.7&75.0 &122.0\\
        &CoTTA  &38.1 &783.6& 39.6 &405.3 &43.4 & 216.2&51.6 &121.6 & 71.4 &74.3 & 91.3 &50.7 &55.9 &275.3\\
        &SWA  & 37.8 &783.6 & 39.1 & 405.3&42.8 &216.2 &50.3 &121.6 & 69.6 & 74.3& 90.1 & 50.7 & 55.0 &275.3\\
        &ECoTTA & 37.2 &174.8 &37.9  &88.8&39.6 &45.8 &46.2 &24.3 & 85.7 &13.6 & 96.4 &8.2 & 57.2 &59.3\\
        &SAR  & \underline{35.5} &367.2 & 36.2 &184.7 &40.2 & 93.5& 68.7& 47.9& 94.1 &25.1 & 98.3 &13.7 & 62.2 &122.0 \\
        &LAW & 35.6 &779.0 & \underline{36.1} &400.7 & \underline{38.1}&211.6& \underline{42.4}&117.0 & 56.9 &69.7 & 92.0 &46.1 &50.2 &270.7\\
        
        &\cellcolor{lightgray}\sysname (Ours) & \cellcolor{lightgray}\textbf{34.3} & \cellcolor{lightgray}88.4&\cellcolor{lightgray}\textbf{34.7} & \cellcolor{lightgray}45.9& \cellcolor{lightgray}\textbf{35.8}&\cellcolor{lightgray}24.7 & \cellcolor{lightgray}\textbf{38.7}& \cellcolor{lightgray}14.1&\cellcolor{lightgray}\textbf{44.5} &\cellcolor{lightgray}8.8 & \cellcolor{lightgray}\underline{73.6} &\cellcolor{lightgray}6.1 & \cellcolor{lightgray}\textbf{43.6} &\cellcolor{lightgray}31.3\\
        \hline

        \multirow{9}{*}{ImageNet-C} 
        & Source  &82.4&1053.2 &82.4 &539.4 & 82.4&282.5 &82.4 &154.0 & 82.4 &89.8 & \textbf{82.4} &57.7 &82.4 & 362.8\\
       & EATA  & \textbf{59.1} &5780.3 & \textbf{60.8} &2903.0 & \underline{63.7} &1464.4 &\underline{68.9} &745.0 & \underline{79.9} &385.4 & 86.4 &205.5 & \underline{69.8} &1913.9\\
        &Continual TENT  & 67.0&5780.3 & 67.5&2903.0 &71.7 &1464.4 &91.7 & 745.0& 97.4&385.4&99.3 &205.5 & 82.4 &1913.9\\
        &CoTTA  &65.9 &11520.4 & 66.1 &5913.5 & 67.3 &3110.1 & 82.2 & 1708.3& 96.5 &1007.5 & 99.7 &657.1 &79.6 &3986.2\\
        &SWA  & 65.3 &11520.4 & 65.8 &5913.5 & 66.7&3110.1 &81.4 & 1708.3& 93.6 &1007.5 & 99.4 &657.1 & 78.7 &3986.2\\
        &ECoTTA &80.8  &2540.2 & 89.7 &1332.7 & 97.8 &729.0 &99.5 &427.2 & 99.8 & 276.2& 99.8 &200.8 &94.6 &917.7 \\
        &SAR  & 61.9 &5780.3 & 62.4 &2903.0 &63.8 &1464.4 &76.7 &745.0 & 80.4 &385.4 & 85.2 &205.5 &71.7 &1913.9\\
        &LAW & \underline{60.7} &11316.0 & \underline{61.4} &5709.1 & \textbf{62.9} &2905.9 & 74.4 &1503.9 & 94.2&803.1 & 98.9 &452.7 & 75.4 &3531.1\\
        
        &\cellcolor{lightgray}\sysname (Ours) & \cellcolor{lightgray}61.4 &\cellcolor{lightgray}1598.4 & \cellcolor{lightgray}62.8 & \cellcolor{lightgray}814.8 & \cellcolor{lightgray}65.1 & \cellcolor{lightgray}423.0 & \cellcolor{lightgray}\textbf{68.1} &\cellcolor{lightgray}227.2 &\cellcolor{lightgray}\textbf{77.0 } &\cellcolor{lightgray}129.2 & \cellcolor{lightgray}\underline{83.0} &\cellcolor{lightgray}80.3 &\cellcolor{lightgray}\textbf{69.6} &\cellcolor{lightgray}545.5 \\
        \bottomrule
        
    \end{tabular}
}
    
    \label{tab:tab4}
\end{table*}

\begin{table*}[t]
\centering
\caption{\textbf{Ablation study on data augmentation type.} From left to right one augmentation type is added at a time.}
\begin{tabular}{lcccc}
\hline
    augmentation type & \textbf{color jitter} & +horizontal flip & +gaussian blur & \textbf{+invert}  \\
    \hline
    Avg err.(\%) & 36.4 & 36.2 & 35.9 & 35.8 \\
    \hline

\end{tabular}

\label{tab:tab5}
\end{table*}

\textbf{Performance with memory constraints.} Evaluating accuracy under memory constraints by adjusting batch sizes to achieve comparable memory consumption among all methods.
Table~\ref{tab:tab2} presents per-domain error rates in CTTA, organized by columns, along with average memory consumption and error rate.
Analyzing Table~\ref{tab:tab2}, we find that \sysname significantly improves accuracy in memory-constrained settings across all datasets and models.
Adapting all BN layer affine parameters (TENT, EATA, SAR) or all model parameters (CoTTA, SWA, LAW) requires storing a considerable amount of activations for backpropagation.
This compels them to use smaller batch sizes to meet memory constraints, leading to significant performance degradation and even collapse.
For example, when the memory constraint is 50MB, TENT achieves 87\% performance on CIFAR100-C, while the original model's performance is 46.8\%. 
Compared to ECoTTA, which requires updating additional side-way meta networks to adapt to the target domain, our \sysname only needs updates for 1.0\% of the representation layers. As a result, it achieves an average accuracy improvement of 14.8\% across three datasets compared to ECoTTA.
Across all models and datasets, \sysname outperforms these state-of-the-art approaches SAR, ECoTTA, and EATA, showing significant accuracy improvements of 4.5\%, 4.9\%, and 14.8\% on CIFAR10-C, CIFAR100-C, and ImageNet-C, respectively.

With these findings, we emphasize \sysname's superiority in resource-constrained environments.
It automatically identifies and adapts a few drift-sensitive representation layers, rather than blindly updating all BN layers.
The shift from BN to representation layers eliminates the need for large batch sizes. Also, by updating adaptation-critical layers only, FoCoTTA avoids storing excessive activations.
This achieves simultaneous optimization of memory efficiency for CTTA adaptation in both batch size and activation. 
Additionally, we uncover challenges faced by full-parameter update methods under memory constraints, particularly performance collapse with smaller batch sizes. 
This research enhances understanding of different methods' adaptability and performance in memory-constrained environments.

\textbf{Performance with the same batch size.} Table \ref{tab:tab4} presents the online measurement average error rate and memory consumption on the corrupted dataset under CTTA, considering different batch sizes, models, and datasets.  Clearly, \sysname achieves a threefold reduction in average memory consumption while improving average accuracy by 8.1\%, 3.6\%, and 0.2\% on CIFAR-10C, CIFAR-100C, and ImageNet-C, respectively. This demonstrates that \sysname maintains robustness across a wide range of batch sizes while remaining memory-efficient. The superior performance of \sysname, from its strategy of updating the adaptation-critical representation layers during adaptation, only mitigates its reliance on batch size. Unlike methods that optimize all BN layers (TENT, EATA, SAR), which rely on larger batch sizes. While methods that update all parameters (CoTTA, SWA, LAW) need to store a large number of activations. \sysname optimizes CTTA from both batch size and activation.  Therefore, \sysname can provide significant memory savings without compromising accuracy.


\begin{table}[t]
\centering
\caption{\textbf{Average inference time per domain on the CIFAR10-C dataset using WideResNet-28 with a batch size of 32.} Also, we show the average error rates (\%) overall 15 corruptions.}
\begin{tabular}{lcc }
\hline
    Method & Adaptation time (s) & Error rate (\%) \\
    \hline
    Source & 5.6 & 43.5\\
    Continual TENT & 33.2 & 46.5\\
    CoTTA & 1854.4  & 22.2\\
    SWA & 1650.6 & 21.7\\
    SAR & 40.7 & 21.4\\
    EATA & 32.1 & 23.9\\
    LAW & 60.2 & 18.6 \\
    ECoTTA & 65.1 & 22.3\\
    \hline
    \textbf{FoCoTTA (Ours)}  & \textbf{16.4} & \textbf{18.9}\\
    \hline

\end{tabular}
\label{tab:taba}
\end{table}

\textbf{Adaptation time.} Table~\ref{tab:taba} presents the average adaptation times across various methods for each domain. These measurements were obtained through testing on the CIFAR10-C dataset using a single NVIDIA 3090 GPU, employing the WideResNet-28 model with a batch size of 32. Also, we show their average error rates overall of 15 corruptions.
From Table.~\ref{tab:taba}, it is clear that \sysname significantly reduces adaptation time with optimal performance compared to state-of-the-art CTTA methods. Specifically, compared to CoTTA, \sysname reduces adaptation time by a factor of 113, while compared to the fastest adaptation method EATA, \sysname improves adaptation time by a factor of 2. This advantage primarily results from \textit{(i)} updating only a few adaptation-critical representation layers, effectively reducing computation during backpropagation, and \textit{(ii)} selecting reliable samples for model optimization, further reducing adaptation time.

\textbf{Integrate BN layer optimization.} In Table~\ref{tab:tabb}, we integrated \sysname with the BN layer optimization method MECTA \cite{hong2023mecta}. Encouragingly, their combined application further reduced \sysname's error rate from 18.9\% to 17.1\%. This not only validates the effectiveness of \sysname but also expands its applicability.

\begin{table*}
\centering
\caption{
\textbf{Integrate \sysname with the BN layer optimization method MECTA.} The evaluation using WideResNet-28 on the CIFAR10-C dataset with a batch size of 32, demonstrates improved performance.}
\resizebox{\linewidth}{!}{
    \begin{tabular}{l|lllllllllllllll|c}
    
        \hline
         \multirow{2}{*}{Method} & \multicolumn{15}{l|}{t $\xrightarrow[\quad\quad
        \quad\quad\quad\quad\quad\quad\quad\quad\quad\quad\quad\quad\quad\quad\quad\quad\quad\quad\quad\quad\quad\quad\quad\quad\quad\quad\quad\quad\quad\quad\quad\quad\quad\quad\quad\quad\quad\quad\quad\quad\quad\quad\quad\quad\quad]{}$} &\\
          & Gaus. & Shot & Impu.&  Defo.&  Glas.&  Moti. & Zoom&  Snow & Fros.&  Fog & Brig.&  Cont.&  Elas.&  Pixe.&  Jpeg&  Avg. err\\

\hline
EATA & 24.9& 21.2 &31.4 & 18.6 & 35.4&21.8& 19.5 &21.9 &20.9& 22.6 & 15.8 &20.3 &30.0&26.1& 28.0&23.9  \\
+ MECTA & 25.8& 20.1 &29.9 & 12.6 & 31.0&13.7 & 11.0 &16.1 &15.1 & 13.3 & 8.3 &11.0&20.0 &15.7 &20.8 &17.6  \\

\hline
TENT & 26.3& 25.8 &38.5 & 32.2 & 48.2&40.2 & 43.7 &47.5 &45.7& 48.1 & 46.2 &62.6 &61.9&63.9& 66.7&46.5  \\
+ MECTA & 25.6& 20.7 &28.9 & 13.2 & 29.7&14.1 & 12.0 &16.8 &15.4 & 15.2 & 8.7 &13.5&19.5 &15.5 &20.3 &17.9  \\

        \hline
\textbf{FoCoTTA (Ours)} & 23.8& 22.3 &30.1 & 13.7 & 32.2&14.4 & 12.6 &16.4 &16.5 & 13.8 & 8.6 &16.5 &24.1 &16.7 & 21.1&18.9  \\
\textbf{+ MECTA} & 22.6& 19.1 &25.2 & 13.0 & 32.1&13.7 & 10.9 &14.9 &15.2 & 12.8 & 7.8 &12.8 &21.7 &15.0 &19.6 &17.1  \\
        \hline
    \end{tabular}
}
\label{tab:tabb}
\end{table*}

\textbf{Comparison with fine-tuning methods.} {We conducted a comparison between \sysname and various methods that directly fine-tune the adaptation model on the CIFAR10-C dataset. The evaluation utilized the WideResNet-28 model, employing the Adam optimizer with a learning rate of 0.001 and a batch size of 32. To ensure a fair comparison, all methods utilized identical hyperparameters throughout the adaptation process.
From Table~\ref{tab:tabc}, we found that fine-tuning all convolutional layers yields the poorest performance, whereas fine-tuning the BN layer results in suboptimal performance. Notably, the exclusive fine-tuning of the BN layer, incorporating tracking statistics during adaptation, leads to a lower error rate. Compared to all methods, our \sysname consistently attains the lowest error rate.

\begin{table*}
\centering
\caption{ \textbf{We conduct a comparison between \sysname and various methods that directly fine-tune the adaptation model on the CIFAR10-C dataset.} The evaluation utilized the WideResNet-28 model with a batch size of 32, and the results demonstrate a substantial improvement in performance with our approach.
}
\resizebox{\linewidth}{!}{
    \begin{tabular}{l|lllllllllllllll|c}
    
        \hline
         \multirow{2}{*}{Method} & \multicolumn{15}{l|}{t $\xrightarrow[\quad\quad
        \quad\quad\quad\quad\quad\quad\quad\quad\quad\quad\quad\quad\quad\quad\quad\quad\quad\quad\quad\quad\quad\quad\quad\quad\quad\quad\quad\quad\quad\quad\quad\quad\quad\quad\quad\quad\quad\quad\quad\quad\quad\quad\quad\quad\quad]{}$} &\\
          & Gaus. & Shot & Impu.&  Defo.&  Glas.&  Moti. & Zoom&  Snow & Fros.&  Fog & Brig.&  Cont.&  Elas.&  Pixe.&  Jpeg&  Avg. err\\
        
        \hline

FC-Tune & 28.9& 27.2 &37.3 & 14.0 & 36.1&15.3 & 13.5 &18.7 &18.8& 16.6 &9.2 &14.2 &25.2 &21.0 &28.2 &21.6  \\
Conv-Tune & 90.1& 89.1 &90.4 & 89.9 & 90.0&87.6& 89.3 &89.6 &89.5& 89.9 &89.5 &91.5 &90.0 &89.2 &89.4 &89.7  \\
BN-Tune & 28.3& 26.1 &36.1 & 13.9 & 35.0&15.0 & 13.3 &18.4 &18.8& 16.2 &9.3&14.1 &24.8 &21.1 &27.4 &21.2  \\
No-Adapt & 72.3& 65.7 &72.9 & 46.9 & 54.3&34.8 & 42.0 &25.1 &41.3&26.0 &9.3 &46.7 &26.6 &58.5 &30.3 &43.5  \\
        \hline
        \textbf{FoCoTTA (Ours)} & 23.8& 22.3 &30.1 & 13.7 & 32.2&14.4 & 12.6 &16.4 &16.5 & 13.8 & 8.6 &16.5 &24.1 &16.7 & 21.1&18.9  \\
        \hline
    \end{tabular}
}
\label{tab:tabc}
\end{table*}

\textbf{Visualization.} In \figref{figa}, we visualize the discriminative capabilities of different methods on the same batch sample in CIFAR10-C. From the figure, we observe that our \sysname, like other methods, exhibits discriminative abilities. This further substantiates the effectiveness of \sysname.

\begin{figure}
\centering
\includegraphics[width=1.0\columnwidth]{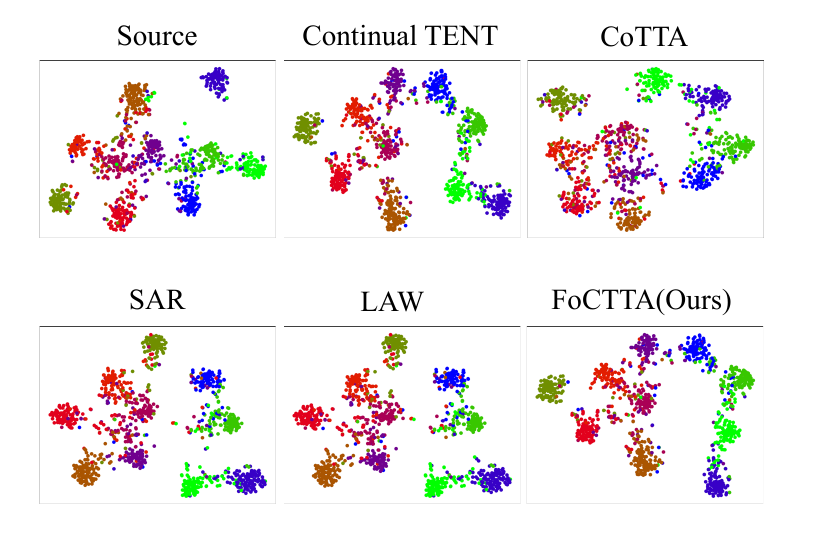}
\caption{\textbf{Visualization of the discriminative power of the sample features of different methods on CIFAR10-C.} Colors represent sample classes.}
\label{figa}
\end{figure}

\begin{table}[t]
\centering
\caption{\textbf{Avg err.(\%) with different values of $\lambda$.}}
\begin{tabular}{llllllll }
\hline
    $\lambda$ & 0.1 & 0.5 & 0.9 & \textbf{1.0} & 1.2 & 1.5 \\
    \hline
    Avg err.(\%) & 36.0 & 35.9 & 35.9 & 35.8 & 36.0 & 36.1 \\
    \hline

\end{tabular}

\label{tab:tab6}
\end{table}

\begin{table}[t]
\centering
\caption{\textbf{Avg err.(\%) with different values of $\alpha$.}}
\begin{tabular}{lllllll}
\hline
    $\alpha$ & 0.03 & 0.05 & 0.08 & \textbf{0.1} & 0.15 & 0.20 \\
    \hline
    Avg err.(\%) & 37.9 & 36.8 & 36.0 & 35.8 & 35.8 & 36.1 \\
    \hline

\end{tabular}

\label{tab:tab7}
\end{table}

\subsection{Ablation Study}
In the following experiments, if not specified, we use CIFAR100-C and robustly pre-trained WideResNet-40.

\textbf{Necessity of Each Design.} Table~\ref{tab:tab3} demonstrates the influence of removing individual designs in \sysname on its performance. The results show a significant performance decline when any \sysname design is removed, highlighting the crucial role of each design in achieving exceptional performance. Particularly, the choice of the adaptation-critical layer is vital for \sysname, and its absence leads to performance collapse.



\textbf{Influence of $\lambda$ and $\alpha$:} Table~\ref{tab:tab6} and Table~\ref{tab:tab7} show the effect of $\lambda$ and $\alpha$ on the error rate of \sysname, respectively. Here, $\lambda$ represents the weight of the regularization term. $\alpha$ represents the number of adaptation-critical layers selected for optimization. We choose the setting that achieves the best performance, with $\lambda$ = 1.0 and $\alpha$ = 0.1 as default values.


\textbf{Influence of the data augmentation type:} We use data augmentation in the warm-up training phase to simulate domain shifts.
Table~\ref{tab:tab5} demonstrates the influence of data augmentation types on \sysname, revealing its robustness.
By using default settings, we choose color jittering and inverting, achieving the best performance.


\section{Conclusion}
\label{conclusion}
In this paper, our focus is on addressing the memory efficiency issues during the adaptation of CTTA. 
To this end,  we propose FoCTTA, a new low-memory CTTA strategy.
Rather than blindly update all BN layers, we identify and adapt a few drift-sensitive representation layers. 
The shift from BN to representation layers eliminates the need for large batch sizes. 
Furthermore, by updating adaptation-critical layers only, FoCTTA avoids storing excessive activations.  This approach improves memory efficiency and maintains adaptation effectiveness, as confirmed by evaluations across various models and datasets.

\newpage
\bibliography{aaai25}


\end{document}